# LLM-assisted Vector Similarity Search

Md Riyadh, Muqi Li, Felix Haryanto Lie, Jia Long Loh, Haotian Mi, Sayam Bohra

*Original article: [https://engineering.grab.com/llm-assisted-vector-similarity-search](https://engineering.grab.com/llm-assisted-vector-similarity-search)*

## Introduction

As the complexity of data retrieval requirements continue to grow, traditional search methods often struggle to provide relevant and accurate results, especially for nuanced or conceptual queries. Vector similarity search has emerged as a powerful technique for finding semantically similar information. It refers to finding vectors in a large dataset that are most similar to a given query vector, typically using some distance or similarity measure. The concept originated in the 1960s with the work by Minsky and Papert on nearest neighbour search [1]. Since then, the idea has evolved substantially with modern approaches often using approximate methods to enable fast search in high-dimensional spaces, such as locality-sensitive hashing [2] and graph-based indexing [3].

Recently, vector similarity search has become a crucial component in many machine learning and information retrieval applications. It is one of the key technologies that popularised the idea of Retrieval Augmented Generation (RAG) [4] which increased the applicability of Transformer [5] based Generative Large Language Models (LLMs) [6] in domain-specific tasks without requiring any further training or fine-tuning. However, the effectiveness of the vector search can be limited when dealing with intricate queries or contextual nuances. For example, from a typical vector similarity search perspective, "I like fishing" and "I do not like fishing" may be quite close to each other, while in reality, they are the exact opposite. In this blog post, we discuss an approach that we experimented with that combines vector similarity search with LLMs to enhance the relevance and accuracy of search results for such complex and nuanced queries. We leverage the strengths of both techniques: vector similarity search for efficient shortlisting of potential matches, and LLMs for their ability to understand natural language queries and rank the shortlisted results based on their contextual relevance.

---

[1] M. Minsky and S. Papert, Perceptrons: An Introduction to Computational Geometry. MIT Press, 1969.
[2] P. Indyk and R. Motwani, "Approximate nearest neighbors: Towards removing the curse of dimensionality," in Proceedings of the Thirtieth Annual ACM Symposium on Theory of Computing, 1998.
[3] Y. Malkov and D. Yashunin, "Efficient and robust approximate nearest neighbor search using hierarchical navigable small world graphs," IEEE Transactions on Pattern Analysis and Machine Intelligence, 2020.
[4] P. Lewis, E. Perez, A. Piktus, F. Petroni, V. Karpukhin, N. Goyal, and D. Kiela, "Retrieval-augmented generation for knowledge-intensive NLP tasks," in Advances in Neural Information Processing Systems, 2020.
[5] A. Vaswani, "Attention is all you need," in Advances in Neural Information Processing Systems, 2017.
[6] A. Radford, "Improving language understanding by generative pre-training," 2018.

## Proposed solution

The proposed solution involves a two-step process:

1. Vector similarity search: We first perform a vector similarity search on the dataset to obtain a shortlist of potential matches (e.g., top 10-50 results) for the given query. This step leverages the efficiency of vector similarity search to quickly narrow down the search space.

2. LLM-assisted ranking: The shortlisted results from the vector similarity search are then fed into an LLM, which ranks the results based on their relevance to the original query. The LLM's ability to understand natural language queries and contextual information helps in identifying the most relevant results from the shortlist.

By combining these two steps, we aim to achieve the best of both worlds: the efficiency of vector similarity search for initial shortlisting, and the contextual understanding and ranking capabilities of LLMs for refining the final results.

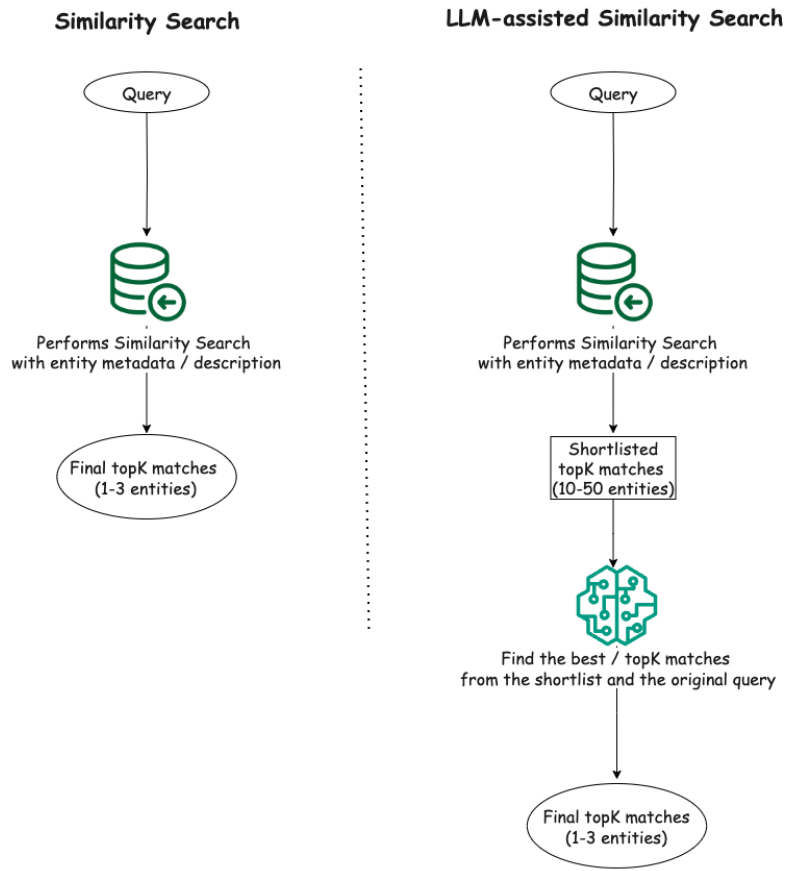

Figure 1. Similarity search and the proposed LLM-assisted similarity search.

# Experiment

## Datasets

To evaluate the effectiveness of our proposed solution, we conducted experiments on two small synthetic datasets in CSV format that we curated using GPT-4o [7].

- **Food dataset**: A collection of 100 dishes with their titles and descriptions.
- **Tourist spots dataset**: A collection of 100 tourist spots in Asia, including their names, cities, countries, and descriptions.

It is important to note that we primarily focus on performing similarity search on structured data such as description of various entities in a relational database.

## Setup

Our experimental setup included a Python script for vector similarity search leveraging Facebook AI Similarity Search (FAISS) [8], a library developed by Facebook that offers efficient similarity search, and OpenAI's embeddings (i.e., text-embedding-ada-002) [9] to generate the vector embeddings needed for facilitating the vector search. For our proposed solution, an LLM component (i.e., GPT-4o) was included in the setup in addition to the FAISS-based similarity search component.

## Observations

To compare the performance of the proposed approach of LLM-assisted vector similarity search as outlined in the "Proposed solution" section with the raw vector similarity search, we conducted both techniques on our two synthetic datasets. With the raw vector search, we get the top three matches for a given query. For our proposed technique, we first get a shortlist of 15 entity matches from FAISS for the same query, and supply the shortlist and the original query to LLM with some descriptive instructions in the prompt to find the top three matches from the provided shortlist.

From the experiments, in simpler cases where the queries were straightforward and directly aligned with the textual content of the data, both the raw similarity search and the LLM-assisted similarity search demonstrated comparable performance. However, as the queries became more complex, involving additional constraints, negations, or conceptual requirements, the LLM-assisted search exhibited a clear advantage over the raw similarity search. The LLM's ability to understand context and capture subtleties in the queries allowed it to filter out irrelevant results and rank the most appropriate ones higher, leading to improved accuracy.

---

[7]"Hello GPT-4o," OpenAI, May 2024. [Online]. Available: https://openai.com/index/hello-gpt-4o/. [Accessed: Oct. 6, 2024].
[8]M. Douze, A. Guzhva, C. Deng, J. Johnson, G. Szilvasy, P. E. Mazaré, and H. Jégou, "The faiss library," arXiv preprint arXiv:2401.08281, 2024.
[9]"Embeddings," OpenAI API. [Online]. Available: https://platform.openai.com/docs/guides/embeddings. [Accessed: Oct. 6, 2024].

Here are a few examples where the LLM-assisted similarity search performed better:

### Food dataset

**Query**: "food with no fish or shrimp"

**Raw similarity search result**:

- title: Tempura, description: A Japanese dish of seafood or vegetables that have been battered and deep fried.

- title: Ceviche, description: A seafood dish popular in Latin America, made from fresh raw fish cured in citrus juices.

- title: Sushi, description: A Japanese dish consisting of vinegared rice accompanied by various ingredients such as seafood and vegetables.

**LLM-assisted similarity search result**:

- title: Chicken Piccata, description: Chicken breasts cooked in a sauce of lemon, butter, and capers.

- title: Chicken Alfredo, description: An Italian-American dish of pasta in a creamy sauce made from butter and Parmesan cheese.

- title: Chicken Satay, description: Grilled chicken skewers served with peanut sauce.

**Observation**: The LLM correctly filtered out dishes containing fish or shrimp, while the raw similarity search failed to do so, presumably due to the presence of negation in the query.

### Tourist spots dataset

**Query**: "exposure to wildlife"

**Raw similarity search result**:

- name: Ocean Park, city: Hong Kong, country: Hong Kong, description: Marine mammal park and oceanarium.

- name: Merlion Park, city: Singapore, country: Singapore, description: Iconic statue with the head of a lion and body of a fish.

- name: Manila Bay, city: Manila, country: Philippines, description: A natural harbor known for its sunset views.

**LLM-assisted similarity search result**:

- name: Ocean Park, city: Hong Kong, country: Hong Kong, description: Marine mammal park and oceanarium.

- name: Chengdu Research Base, city: Chengdu, country: China, description: A research center for giant panda breeding.

- name: Mount Hua, city: Shaanxi, country: China, description: Mountain known for its dangerous hiking trails.

**Observation**: Two out of the top three matches by the LLM-assisted technique seem relevant to the query while only one result from the raw similarity search is relevant and the other two being somewhat irrelevant to the query. The LLM identified the relevance of a research base for giant panda breeding to the "exposure to wildlife", which the raw similarity search ignored in its ranking.

These examples provide a glimpse into the utility of LLMs in finding more relevant matches in scenarios where the queries involved additional context, constraints, or conceptual requirements beyond simple keyword matching. On the other hand, when the queries were more straightforward and focused on specific keywords or phrases present in the data, both approaches demonstrated comparable performance. For instance, queries like "Japanese food" or "beautiful mountains" yielded similar results from both the raw similarity search and the proposed LLM-assisted approach.

Overall, the LLM-assisted vector search exhibited a clear advantage in handling complex queries, leveraging its ability to understand natural language and contextual information. However, for simpler queries, the raw similarity search remained a viable option, especially when computational efficiency is a concern.

# Conclusion

The experiments demonstrated the potential of combining vector similarity search with LLMs to enhance the relevance and accuracy of search results, particularly for complex and nuanced queries. While vector similarity search alone can provide reasonable results for straightforward queries, the LLM-assisted approach shines when dealing with queries that require a deeper understanding of context, nuances, and conceptual relationships. By leveraging the natural language understanding capabilities of LLMs, this approach can better capture the intent behind complex queries and provide more relevant search results.

Our experiment was limited to using a small volume of structured data (100 data points in each dataset) with a limited number of queries. However, we have witnessed similar enhancement in search result relevance when we deployed this solution internally within Grab for larger datasets, for example, 4500+ rows of data stored in a relational database.

Nevertheless, it is important to note that the effectiveness of this approach may still depend on the quality and complexity of the data, as well as the specific use case and query patterns. We believe it is still worthwhile to evaluate the proposed approach for more diverse (e.g., beyond CSV) and larger datasets. An interesting future work can be varying the size of the shortlist from the similarity search and observing how it impacts the overall search relevance when using the proposed approach. In addition, for real world applications, the performance implications in terms of additional latency introduced by the additional LLM query must also be considered.